\documentclass[conference]{IEEEtran}
\IEEEoverridecommandlockouts

\usepackage{amsmath,amssymb,amsthm}
\usepackage{graphicx}
\usepackage{booktabs}
\usepackage{multirow}
\usepackage{url}
\usepackage[most]{tcolorbox}
\usepackage{xcolor}
\usepackage{hyperref}
\usepackage{algorithm}
\usepackage{algorithmic}
\usepackage{balance}
\usepackage{wrapfig}
\usepackage{subcaption}
\usepackage{makecell} 

\usepackage[numbers,sort&compress]{natbib}



\newtcolorbox[auto counter, number within=section]{insight}[1][]{
  colback=black!5,
  colframe=black!60,
  fonttitle=\bfseries,
  title=Insight~\thetcbcounter,
  boxrule=0.5pt,
  arc=3pt,
  left=6pt, right=6pt, top=6pt, bottom=6pt,
  #1
}

\title{\LARGE \bf RACER: Residual-Adaptive Closed-Loop Estimation for Sampling-Based Planning
       in Wheeled-Quadruped Racing}

\author{
Yuxiang Liu$^{1,*}$, Marla Eisman$^{1,*}$, Lizhi Yang$^{2,*}$, Aaron Ames$^{2}$, Francesco Borrelli$^{1}$%
\thanks{$^{*}$Equal contribution.}%
\thanks{$^{1}$University of California, Berkeley, CA, USA.}%
\thanks{$^{2}$California Institute of Technology, Pasadena, CA, USA.}%
}

\begin{document}
\maketitle

\begin{abstract}
We present RACER, a hierarchical control framework for wheel-based quadruped racing that combines an MPPI planner with a learned residual dynamics model and a low-level RL velocity tracker. The planner augments a nominal unicycle kinematic model with a neural residual term to capture the closed-loop tracking behavior of the RL policy. To train this residual model under limited real-world data, we propose Low-Rank Residual Adaptation (LoRRA), a two-stage approach that pre-trains on large-scale simulation data for broad coverage and then fine-tunes on a small real-world dataset with a low-rank constraint. In simulation, we empirically validate our engineering choices by showing (A) Residual dynamics improve the overall performance of our pipeline by capturing the tracking error of RL velocity tracker at high-speed cornering. (B) Residual dynamics trained with both source-domain and target-domain data gives racing performance significantly better than the residual dynamics trained with only target-domain data. (C) Low-rank constraint at target-domain adaptation gives higher success rates and higher performance than full-tune and from-scratch when domain gap in ground coefficient or joint gain increases.
\end{abstract}


\section{Introduction}
\label{sec:intro}

Racing a wheeled quadruped is challenging. First, wheeled quadruped inherits the high-dimensional dynamics of a quadruped with the nonlinear tire dynamics of a racing car ~\citep{bjelonic2020rolling, lee2024wheeledlegged}. This complexity challenges many classical model-predictive control (MPC) approaches: optimizing long-horizon trajectories over such a high-dimensional and non-linear system is challenging ~\cite{katayama2023mpc, bjelonic2022wbmpc}. At the same time, long-horizon is a must in the context of racing ~\cite{rosolia2020race}. This long-horizon requirement makes naive end-to-end reinforcement learning (RL) approaches equally impractical ~\cite{Kirk_2023}. A long planning horizon creates an extremely sparse reward landscape that destabilizes RL training, and even if clever curriculum or reward-shaping techniques succeed in training an agent, the resulting policy rarely generalizes beyond its training distribution.

To handle the system complexity and long-horizon nature of the racing task, we adapt hierarchical control. An RL policy can handle the nonlinear, high-dimensional dynamics of the wheeled quadruped over a short horizon. It is trained to track velocity commands. It abstracts the full joint-space dynamics into a compact, low-dimensional interface of linear and angular velocity~\citep{hwangbo2019agile, rudin2022walk, lee2024wheeledlegged}. A sampling-based planner ~\citep{williams2017information, xue2024dialmpc}, can then operate over this interface to plan trajectories across a long horizon.

While this hierarchical architecture is conceptually sound, a naive implementation will fail at high-velocity(Sec. ~\ref{sec:exp:residual}). We find that if the planner uses simple reduced-order models to propose trajectories~\citep{bjelonic2020rolling, bjelonic2022wbmpc}, the resulting commands will systematically diverge from the robot's true behavior, because velocity tracking is always imperfect: hardware latency, actuator saturation, and tire slip all introduce tracking errors that a kinematic model cannot predict. To bridge this gap, we augment the reduced-order model (namely the unicycle model) with a learned residual dynamics network that compensates for the difference between the analytic kinematic prediction and the actual closed-loop next state of the RL tracker~\citep{roth2025fdm, kim2022fdm}. Training such a residual model, however, is highly nontrivial since it requires both broad data coverage across the state-action space and high data quality that  reflects real-world dynamics. Simulation can provide coverage but suffers from the sim-to-real gap~\citep{tobin2017dr, kumar2021rma}; real-world data provides fidelity but is scarce and expensive to collect on a fragile racing platform. 

To this end, we propose Low-Rank Residual Adaptation (LoRRA), a two-stage training pipeline that first pre-trains the residual network on a large simulated dataset for coverage($10, 000,000$ state transitions). By coverage we mean that these transitions span the range of velocities, yaw rates, and commands the planner queries during high-speed rollouts, so the residual is accurate across the states DIAL-MPC actually visits rather than only near a handful of collected trajectories. Then, we fine-tune it on a small real-world dataset(about $400$ state transitions) under a low-rank parameter constraint that prevents catastrophic forgetting of the simulation prior~\citep{hu2022lora, daneshmand2026slowrl}.
\begin{figure}[t]
    \centering
    \includegraphics[width=1.0\linewidth]{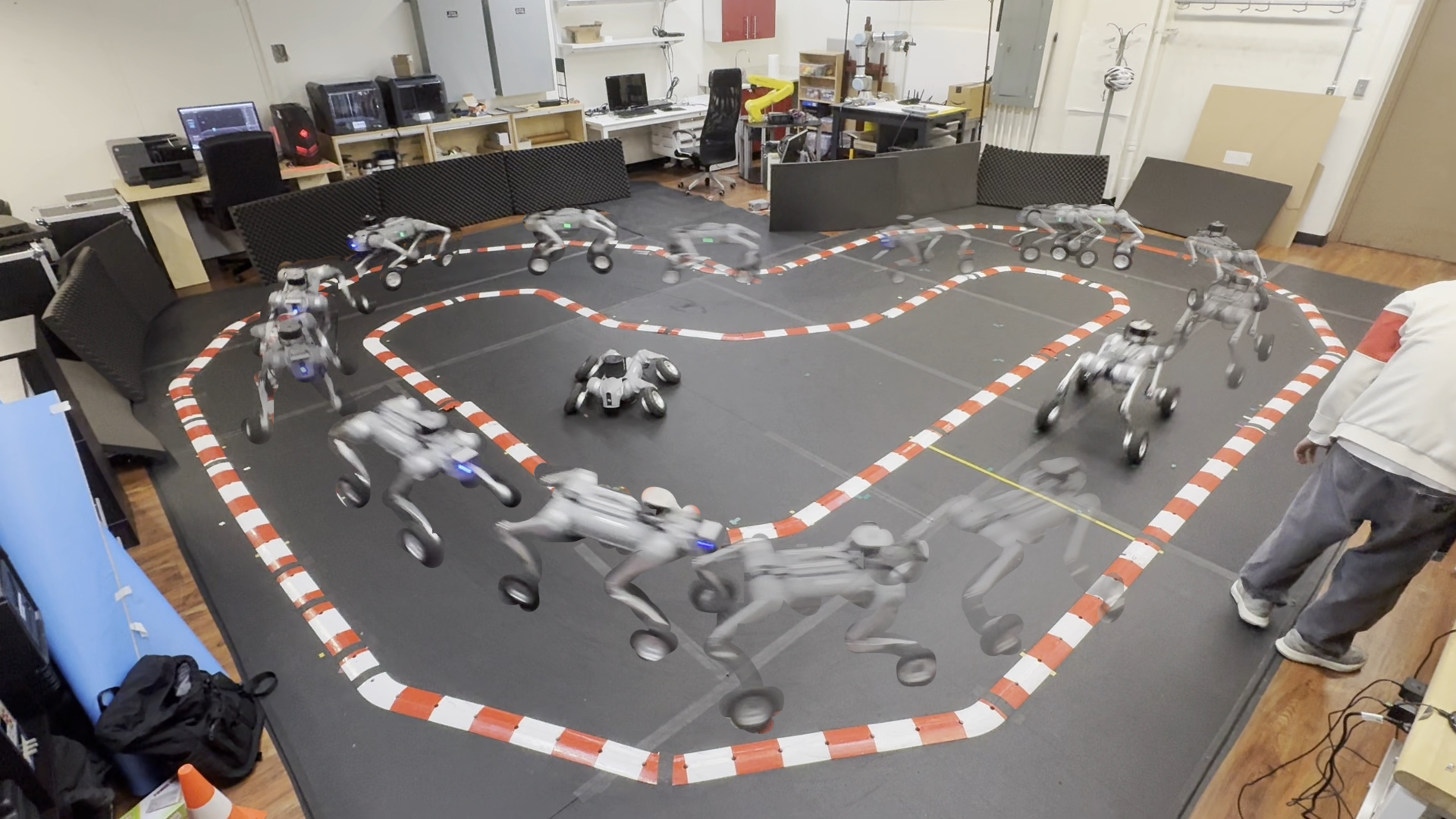}
    \label{fig:front_page}
    \caption{\footnotesize A Go2W Wheeled Quadruped racing along L-shape track using RACER framework. The low-level controller is a velocity RL-tracking policy; the high level planner is a DIAL-MPC controller equipped with residual dynamics trained with LoRRA.}
\end{figure}
Our contribution are as follows:

\begin{itemize}
    \item We propose Residual-Adaptive Closed-Loop Estimation for Sampling-Based Planning
       in Wheeled-Quadruped Racing (RACER), a hierarchical control stack for wheeled-quadruped racing. 
    
    \item We propose LoRRA, a simple yet effective training pipeline for residual dynamics that overcomes the sim-to-real gap.
    
    \item We conduct exhaustive empirical experiments to validate the functionalities of residual dynamics and LoRRA.
\end{itemize}

\section{Related Work}
\label{sec:related}

\paragraph{Wheeled-legged locomotion and racing.}
\citet{bjelonic2020rolling} introduce online trajectory optimization for hybrid walking--driving locomotion on ANYmal-on-wheels, and extend it to whole-body MPC with single-rigid-body dynamics.
\citep{bjelonic2022wbmpc}.
\citet{lee2024wheeledlegged} take a complementary learning-heavy direction, integrating RL locomotion, learned high-level navigation, and large-scale path planning for kilometer-scale urban autonomy. These works mainly focus on hybrid locomotion and urban traveling while high-speed racing with wheeled quadruped is still an engineering setting largely underexplored. The only work to directly address wheeled-quadruped racing ~\cite{eisman2026racing} approaches it from a vehicle-dynamics standpoint. In contrast, we focus on functionalities of residual dynamics. ~\cite{eisman2026racing} and our approach serve as complementary roles.

\paragraph{Learned and residual dynamics in sampling-based MPC.}
Sampling-based MPC, and MPPI in particular~\citep{williams2017information}, is a natural consumer of learned dynamics because it only requires forward simulation.
\citet{roth2025fdm} learn a perceptive forward dynamics model on ANYmal and roll it out inside MPPI for safe navigation, reporting a 41\,\% improvement in position prediction; \citet{kim2022fdm} pair a learned quadruped forward model with an informed trajectory sampler for safe local planning.
On the ground-vehicle side, \citet{kato2026mppipid} fuse a physical model with a neural residual and use MPPI to optimize PID gains for a mini-forklift, and \citet{song2026icode} learn a continuous-time residual via input-concomitant Neural ODEs.
\citet{pezzato2025sampling} sidestep model learning by treating IsaacGym itself as the rollout simulator, while \citet{xue2024dialmpc} push sampling MPC to torque-level quadruped control using full-order dynamics and a diffusion-style annealing schedule that we adopt for our planner.
Relative to this body of work, our residual is distinctive in two ways. First, it is parameterized at the planner's \emph{command interface} (rather than at the full robot state); second, its training pipeline is dual-phased, consuming both simulation data and hardware data.

\begin{figure*}[t]
  \centering
  \includegraphics[width=\linewidth]{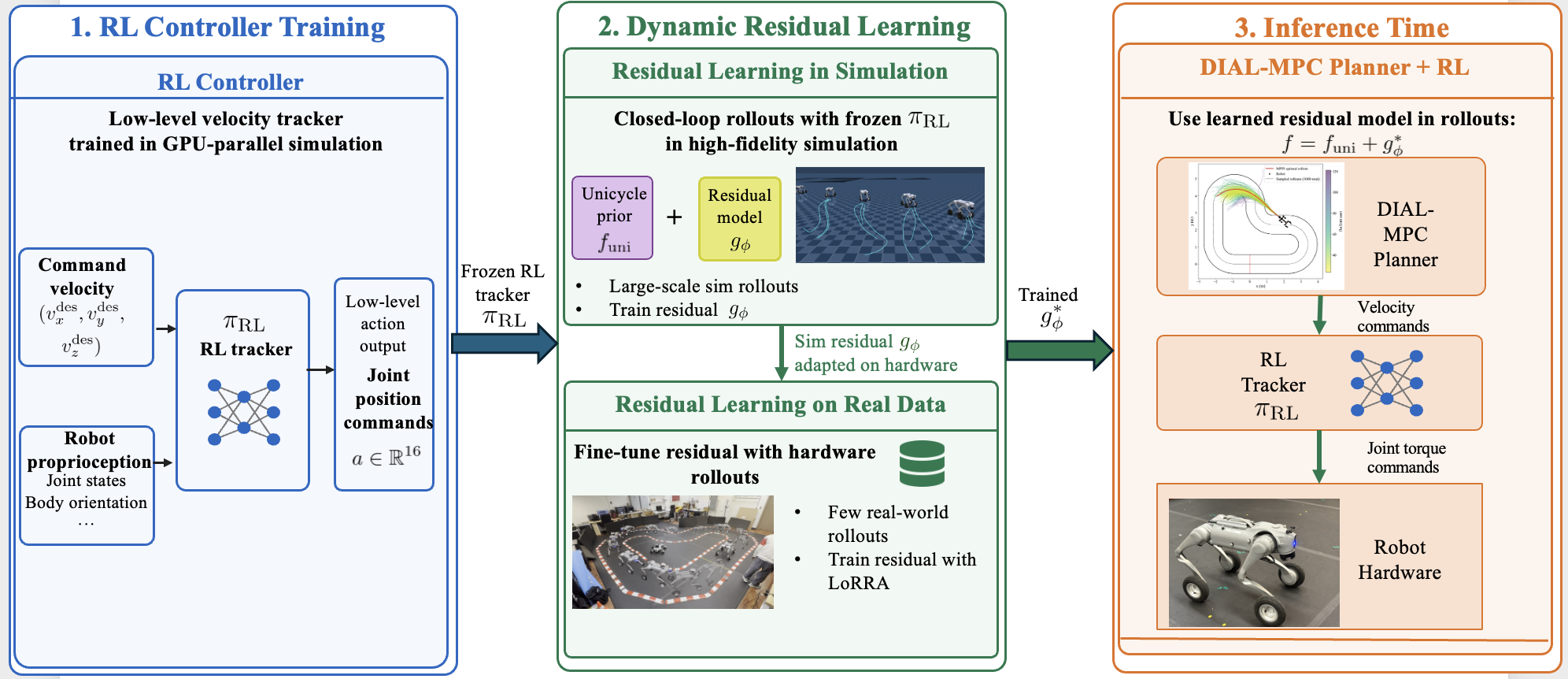}
  \caption{\footnotesize  \textbf{System overview.}
    An DIAL-MPC planner samples velocity commands at 10\,Hz and rolls them out under $f = f_{\mathrm{uni}} + g_{\phi^*}$.
    An RL tracker realises the chosen command at 50\,Hz, producing 16-D joint targets for the Go2-W's 12 leg joints and 4 wheel motors.
    Hardware adaptation refines \emph{only} $g_{\phi^*}$ to compensate domain shift, leaving the planner and tracker untouched.}
  \label{fig:stack}
\end{figure*}

\paragraph{Sim-to-real adaptation of dynamics models.}
\citet{heiden2021neuralsim} augment a differentiable rigid-body simulator with neural networks that absorb unmodeled contact effects; \citet{cheng2022difftune} tune controller parameters through auto-differentiation, and \citet{krishna2025diffcotune} co-tune controller and simulator parameters from a few deployment rollouts.
\citet{he2025asap} learn a \emph{delta-action} residual on a humanoid to realign IsaacGym with the deployed robot; \citet{levy2024ssrl} learn semi-structured dynamics from three minutes of real Go1 data to bootstrap model-based RL from scratch. These works exclusively focus on the sim-to-real of policy, and to the best of our knowledge, we are the first work that studies the sim-to-real transfer of residual dynamics. ~\cite{lee2025sym2real} is the closest work. ~\cite{lee2025sym2real} freezes a symbolic-regression base learned in low-fidelity simulation and fits a fresh neural residual per target domain, at the level of open-loop plant dynamics queried directly by MPC. In contrast, RACER (i) parameterizes its residual at the planner's command interface to model the closed-loop tracking error of a learned RL velocity tracker, and (ii) transfers the pretrained residual itself across the domain gap under a low-rank constraint, rather than re-fitting a residual from scratch.

\paragraph{Parameter-efficient fine-tuning and adaptation regimes.}
Low-Rank Adaptation~\citep{hu2022lora} freezes pre-trained weights and trains low-rank updates that can be merged with no runtime cost; in robot learning, \citet{daneshmand2026slowrl} combine LoRA with a recovery-policy safety mechanism to fine-tune a Unitree Go2 locomotion \emph{policy}, reporting 46.5\,\% lower fine-tuning time and rank-1 adapters sufficing to recover pre-trained real-world performance. To the best of our knowledge, we are the first work that investigates the impacts of low rank constraint on the training of residual dynamics rather than policy adaptation.



\section{Hierarchy Overview: RACER}
\label{sec:method}
RACER employs a two-level hierarchical control architecture: (1) a DIAL-MPC planner that computes optimal velocity commands(More details about the planner's cost function in Appendix Sec. ~\ref{app:mppi:global}), and (2) an RL velocity tracker that executes these commands (Sec.~\ref{sec:method:tracker}).

Our primary contribution lies in the planner layer. To roll out trajectories, the planner uses a discrete-time planar kinematic model augmented with residual dynamics:
\begin{equation}
    s_{t+1} = f_{\mathrm{uni}}(s_t, a_t) + g_\phi(s_t, a_t),
    \label{eq:residual_decomp}
\end{equation}
where the state $s_t = [x_t, y_t, \psi_t, v_{x,t}, \omega_{z,t}]^\top$ comprises the quadruped's base-link world-frame position, yaw, body-frame longitudinal velocity, and yaw rate. The action $a_t = [v_{x,t}^{\mathrm{des}}, \omega_{z,t}^{\mathrm{des}}]^\top$ represents the desired velocity command passed to the low-level RL tracker. The term $f_{\mathrm{uni}}$ denotes a nominal kinematic unicycle model, while $g_\phi$ is the learned residual dynamics. Equation~\ref{eq:residual_decomp} models the closed-loop behavior of the RL tracker, predicting the resulting next state $s_{t+1}$ given the current state $s_t$ and the commanded action $a_t$. In this work, $g_{\phi} $ is implemented as a 5-layer MLP sharing the same input and output space as $f_{\text{uni}}$.

\subsection{Low-Level RL Velocity Tracker}
\label{sec:method:tracker}

The low level RL velocity tracker performs joint-space locomotion control to track the velocity command given by the planner. Specifically, it is a feed-forward neural policy mapping a proprioceptive observation (joint angle, joint velocity, base-link pose) and the planner velocity command to a 16-D joint-actuator output: 12 leg-joint position targets and 4 wheel-velocity targets. Its training follows the standard PPO paradigm ~\cite{fanziqi2024robotlab}. Remarkably, the reward function of RL training is specifically designed to encourage RL tracker to behave closely to $f_{uni}$. Thus, the residual dynamics can be interpreted as an estimation of tracking error of this RL policy, bridging the gap between ideal kinematic commands and the inevitable imperfections of real-world execution caused by hardware limitations. 

\subsection{Training Residual Dynamics}
\label{sec:method:prism}
\label{sec:method:learn}
$g_{\phi}$ is trained with two stages: a simulated pretraining stage on a source-domain dataset $\mathcal{D}_s$ that provides data coverage, followed by an on-robot refinement stage on a small target-domain dataset $\mathcal{D}_t, \; \text{where} \; \lvert \mathcal{D}_s\rvert \gg \lvert\mathcal{D}_t\rvert$ with a Low Rank constraint. In sim-to-real transfer literature~\cite{8460528} \cite{tobin2017dr}, the \emph{source domain} refers to the accessible training environment (here, simulation), the \emph{target domain} is the final deployment environment (here, real-world quadruped hardware deployment), and the \emph{domain gap} is the discrepancy in transition dynamics between the two.

Both stages minimize the residual MSE loss
\begin{equation}
  \mathcal{L}(\phi; \mathcal{D})
  \;=\;
  \mathbb{E}_{(s, a, s') \sim \mathcal{D}}\,
    \bigl\|\, g_\phi(s, a)
              \,-\, \bigl(s' - f_{\mathrm{uni}}(s, a)\bigr) \bigr\|_2^2,
  \label{eq:residual_loss}
\end{equation}

For the rest of this paper, we will refer to our residual dynamics training recipe as \textbf{LoRRA}, Low-rank Residual Adaptation.

Standard class-balancing techniques for single-stage joint training fail due to the severe dataset imbalance ($\lvert \mathcal{D}_s\rvert \gg \lvert\mathcal{D}_t\rvert$) ~\cite{Buda_2018, byrd2019effectimportanceweightingdeep}. Naively re-weighting the target loss induces conflicting gradients: when simulation and hardware data present conflicting physical outcomes for similar states, the network compromises by learning an inaccurate average dynamic ~\cite{yu2020gradientsurgerymultitasklearning, liu2024conflictaversegradientdescentmultitask}.

\paragraph{Low-Rank Adaptation} Our use of low-rank constraints during target-domain adaptation is based on current large language model (LLM) methods ~\cite{biderman2024loralearnsforgets, shuttleworth2025loravsfinetuningillusion, wang2023orthogonalsubspacelearninglanguage}. We constraint the degrees being updated during $\mathcal{D}_t$ adaptation to be less than rank-$r$. The critical challenge in this dual-phase training is catastrophic forgetting. If we perform full-parameter fine-tuning on the small target-domain dataset $\mathcal{D}_t$, the network will catastrophically forget about the posterior learned from the simulation dataset $\mathcal{D}_s$. Previous works have widely validated that low-rank constraint during post-training can keep fine-tuned LLMs closer to the pre-training prior ~\cite{biderman2024loralearnsforgets, shuttleworth2025loravsfinetuningillusion, wang2023orthogonalsubspacelearninglanguage}. In the context of residual dynamics, the low-rank constraint freezes the pre-trained base weights and confines the target-domain updates to a restricted rank-$r$ subspace. This ensures that the network can absorb hardware-specific corrections without forgetting priors learned from $\mathcal{D}_s$.

More details about low-rank constraint is in Appendix. \ref{App:LoRRA_details}.

\begin{figure}[t]
    \centering
    \includegraphics[width=1.0\linewidth]{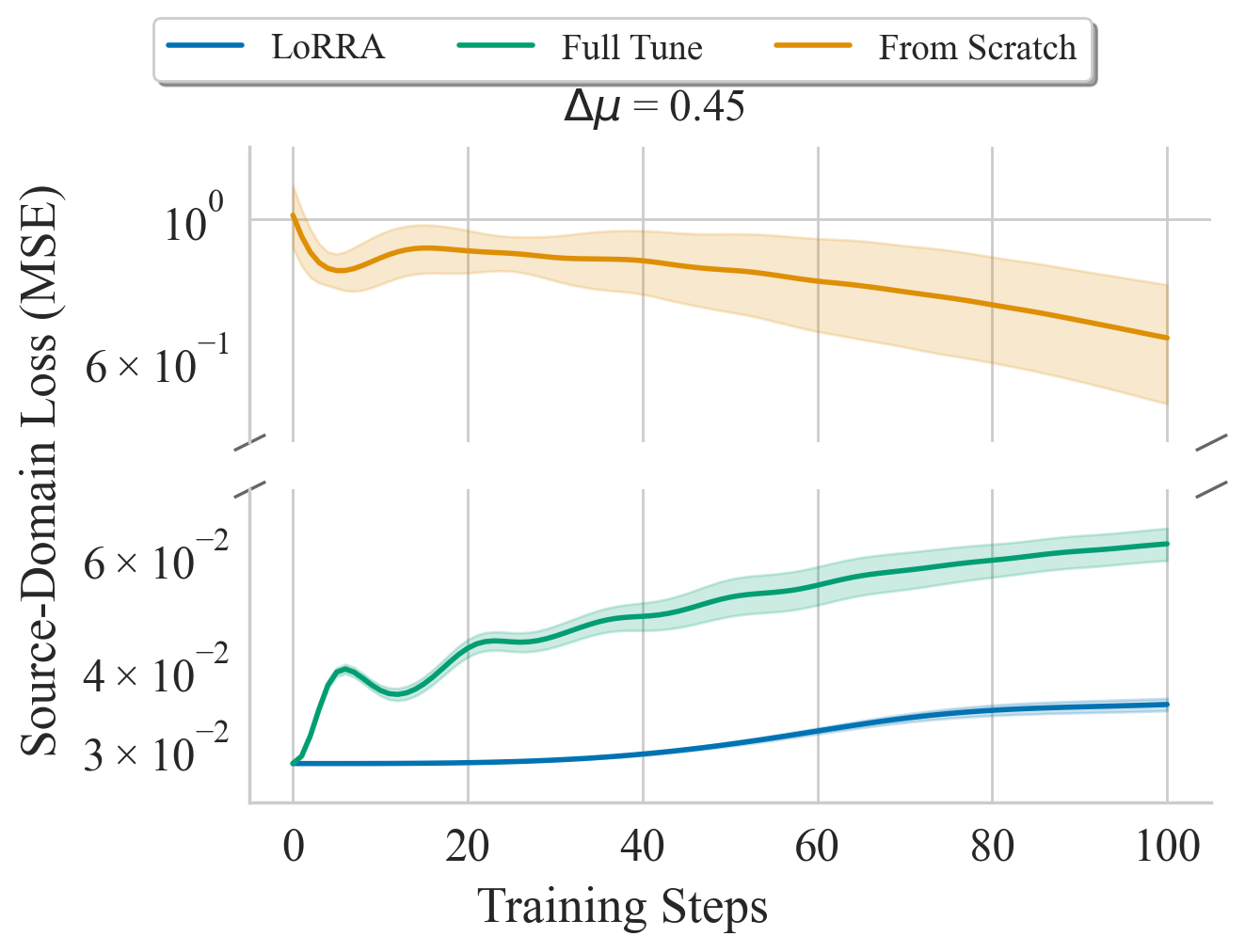}
    \caption{\footnotesize  The Source-Domain loss of residual dynamics $\mathcal{L}(\phi, \mathcal{D}_s)$ through the training with target-domain data $\mathcal{D}_t$. See Sec. ~\ref{sec:exp:lowrank} for definition of each training recipe: LoRRA, Full Tune, and From Scratch. Source and target environments are defined to be two simulation dynamics with different tire frictions, $|\mu| = 0.4$. See Sec. ~\ref{sec:exp:low_rank_sim} for more details about domain definition. \textbf{LoRRA(dual-phase training with low-rank constraint) helps preserve the most source-domain posterior with lowest source-domain loss after target-domain adaptation.}} 
    \label{fig:lorra_train_source_loss_mu0.40}
    \vspace{-10pt}
\end{figure}


\section{Experiments}
\label{sec:exp}
The experiments validate our two main methods: residual dynamics and LoRRA. More specifically, we verify the following claims: \begin{itemize}
\item The proposed residual dynamics model improves racing performance of of a wheeled-legged robot (Sec. ~\ref{sec:exp:residual}).
\item The proposed dual-phase training, LoRRA, improves racing performance of a wheeled-legged robot under large domain gaps.(Sec.~\ref{sec:exp:lowrank}).
\end{itemize}

Following the autonomous racing convention~\citep{rosolia2020race}, we report
three metrics to quantify racing performance. (1) velocity averaged across laps. A lap's velocity will be punished to $0$ if undesirable contacts happen during the lap  (2) maximum deviation from the reference trajectory(Appendix Sec. ~\ref{app:mppi:global}). (3) the number of collision-free laps (free of collisions
and track-boundary violations)completed within a fixed time.

\subsection{Ablation Study of Residual Dynamics}
\label{sec:exp:residual}
In this section, we define naive unicycle to be the variant of RACER with residual dynamics ablated. The naive unicycle model is equivalent to RACER except that the naive unicycle's planner does not use residual dynamics. 

\begin{insight}[label=ins:residual_dynamics]
Residual dynamics improve the overall performance of our pipeline by capturing the tracking error of RL velocity tracker at high-speed cornering.
\end{insight}

\begin{figure}[t]
    \centering
    \includegraphics[width=\linewidth]{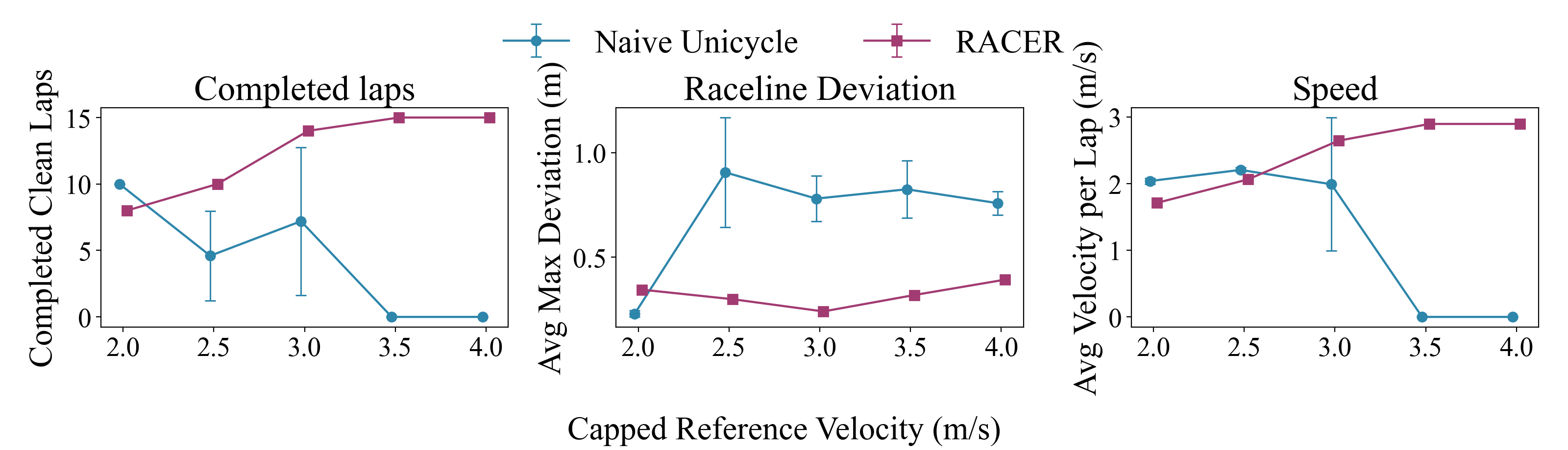}
    \caption{\footnotesize  Simulation performance of Naive Unicycle (Blue) and RACER (Red) v.s. Capped velocity. \textbf{RACER has stable performances independent of the velocity cap while the performance of naive unicycle is downgraded as the cap becomes higher.} Horizontal axis: cap velocity. Vertical Axis: the performance metrics. See Sec. ~\ref{sec:exp:residual} for the definition of naive unicycle. See Sec. ~\ref{sec:exp:residual:sim} for interpretations.}
    \label{fig:nominal_speed_sweep}
\end{figure}

\begin{figure}[t]
    \centering
    \includegraphics[width=\linewidth]{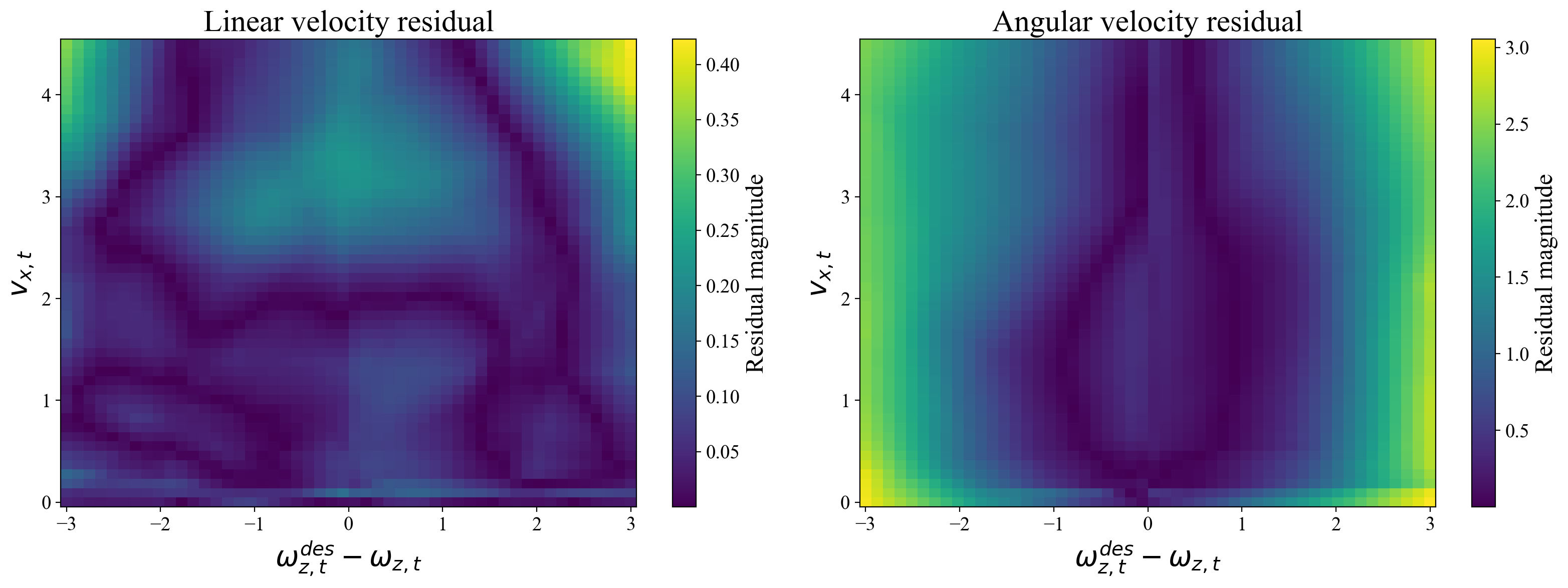}
    \caption{\footnotesize 
    Residual Magnitude under different linear velocities(horizontal axis) and yaw rate changes(vertical axis). \textbf{Residual magnitude positively correlates with linear velocities and yaw rate changes.} Left: Residual of linear velocity; Right: residual of angular velocity. The brighter the area, the larger the residual.}
    \label{fig:residual_magnitude_ablation}
\end{figure}

\subsubsection{Simulation Results}
\label{sec:exp:residual:sim}
  In simulation, we enforce a range of velocity caps of the format $v_x \leq c$, where $c$ is a constant threshold to the planners. 
  The performance metrics of naive unicycle and RACER versus the magnitude of the velocity caps are reported in Fig. ~\ref{fig:nominal_speed_sweep}. This gives us three observations:   i) RACER has low-variance performances regardless of the velocity caps;
  ii) if we enforce a sufficiently conservative velocity cap to naive unicycle, it can finish clean laps;
  iii) and as the velocity cap becomes higher, the performance of naive unicycle can be shown downgraded with worse tracking accuracy, higher-chances of failing laps, and lower average velocity. 
  

We propose the following explanation to the observations:
\begin{itemize}
\item The RL velocity tracker can accurately track the unicycle nominal model when the velocity is low.
\item The tracking error of the RL velocity tracker explodes at high-velocity cornering. The residual dynamics informs RACER about such errors while naive unicycle is blind to the errors.
\end{itemize}

\begin{figure}[t]
    \centering
    \includegraphics[width=\linewidth]{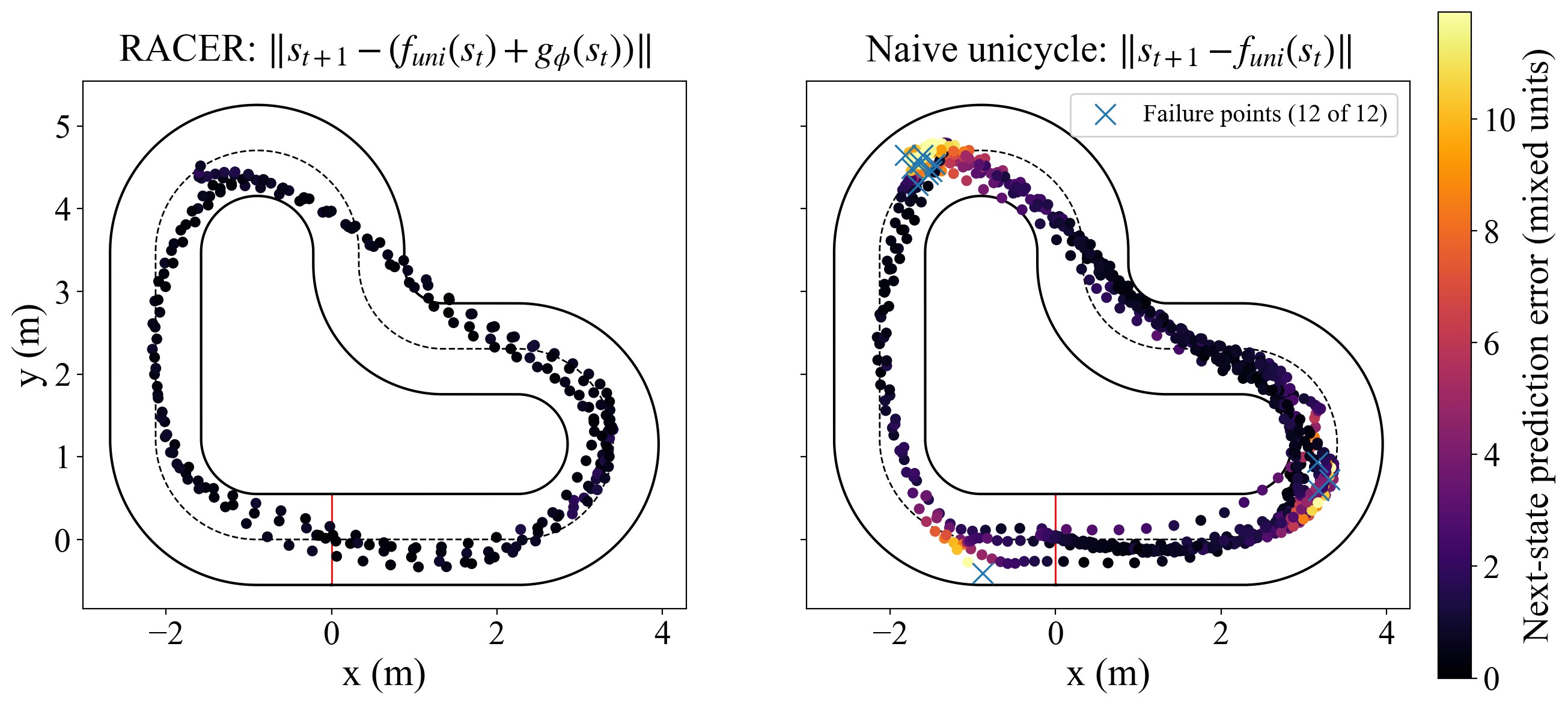}
    \caption{\footnotesize 
    Next-State Prediction Errors of RACER(left) and Naive Unicycle(right). \textbf{Naive Unicycle's prediction errors explode while cornering while RACER keeps the next-state prediction errors consistenly low.} Heat Bar: next-state prediction errors. The brighter the points, the larger the prediction errors. Left: $\| s_{t+1} - (f_{\text{uni}}(s_t) + g_{\phi}(s_t))\|$. Right: $\| s_{t+1} - f_{\text{uni}}(s_t)\|$. The trajectories of left are collected using RACER; The trajectories of right are collected using naive unicycle. A cross dot represents a lap failure.}
    \label{fig:residual_track_geometry}
\end{figure}

We support our explanation with Fig. ~\ref{fig:residual_track_geometry} and Fig. ~\ref{fig:residual_magnitude_ablation}. Fig. ~\ref{fig:residual_track_geometry} shows that RACER has consistently-low next-state predictions while Naive Unicycle's predictions errors explode at high-velocity cornering. Per fig. \ref{fig:residual_magnitude_ablation}, we present the following observations:

 \begin{itemize}
\item Tracking error generally positively correlates with increases in linear velocity and yaw rate.
\item The relationship between the RL agent's tracking error, linear velocity, and yaw rate change is highly non-linear.
\end{itemize}

All observations above validate our claim about the failure of naive unicycle and outlines the role of residual dynamics $g_\phi$: capture the highly-nonlinear tracking error of RL velocity tracker, especially at high-speed cornering. 

\subsubsection{Hardware Results}
\label{sec:exp:residual:hardware}

\begin{table}[t]
\centering
\footnotesize
\setlength{\tabcolsep}{5pt}
\renewcommand{\arraystretch}{1.15}
\caption{\footnotesize  \textbf{Hardware counterpart of Fig.~\ref{fig:nominal_speed_sweep}.} Racing metrics (mean~$\pm$~std over three runs) on the physical robot at different velocity caps $v_x$. \textbf{RACER completes every lap and scales its speed as the caps become higher, whereas the naive unicycle finishes laps only at the conservative $1.5$\,m/s cap and fails entirely ($0$ laps) at higher speeds.} Clean laps are out of $3$ per run; \textbf{bold} marks the better value per metric.}
\label{tab:hardware_residual}
\begin{tabular}{@{}c l c c c@{}}
\toprule
velocity cap & \multirow{2}{*}{Method} & Clean laps & Avg. vel. & Max dev. \\
(m/s) & & (of 3) & (m/s) & (m) \\
\midrule
\multirow{2}{*}{1.5} & RACER          & $\mathbf{3.0 \pm 0.0}$ & $1.21 \pm 0.01$          & $\mathbf{0.42 \pm 0.13}$ \\
                     & Naive Unicycle & $2.3 \pm 1.2$          & $\mathbf{1.49 \pm 0.04}$ & $0.59 \pm 0.12$ \\
\midrule
\multirow{2}{*}{2.0} & RACER          & $\mathbf{3.0 \pm 0.0}$ & $\mathbf{1.41 \pm 0.02}$ & $\mathbf{0.45 \pm 0.08}$ \\
                     & Naive Unicycle & $0.0 \pm 0.0$          & $0.00$                   & $0.83 \pm 0.57$ \\
\midrule
\multirow{2}{*}{3.2} & RACER          & $\mathbf{3.0 \pm 0.0}$ & $\mathbf{1.95 \pm 0.04}$ & $\mathbf{0.35 \pm 0.03}$ \\
                     & Naive Unicycle & $0.0 \pm 0.0$          & $0.00$                   & $0.80 \pm 0.22$ \\
\bottomrule
\end{tabular}
\end{table}

On the physical robot we repeat the velocity sweep of Sec.~\ref{sec:exp:residual:sim}, comparing RACER against the naive unicycle at velocity caps $c \in \{1.5, 2.0, 3.2\}$\,m/s with three runs each. Table~\ref{tab:hardware_residual} reports the same three metrics as Fig.~\ref{fig:nominal_speed_sweep} and reproduces the simulation pattern. At the conservative $1.5$\,m/s cap the naive unicycle is even able to finish laps at a slightly higher average velocity. As the cap increases naive unicycle fails to complete a single lap, whereas RACER completes all laps, maintains the lowest raceline deviation, and keeps raising its average velocity. This confirms Insight~\ref{ins:residual_dynamics} on hardware.

\begin{figure}[t]
    \centering
    \includegraphics[width=\linewidth]{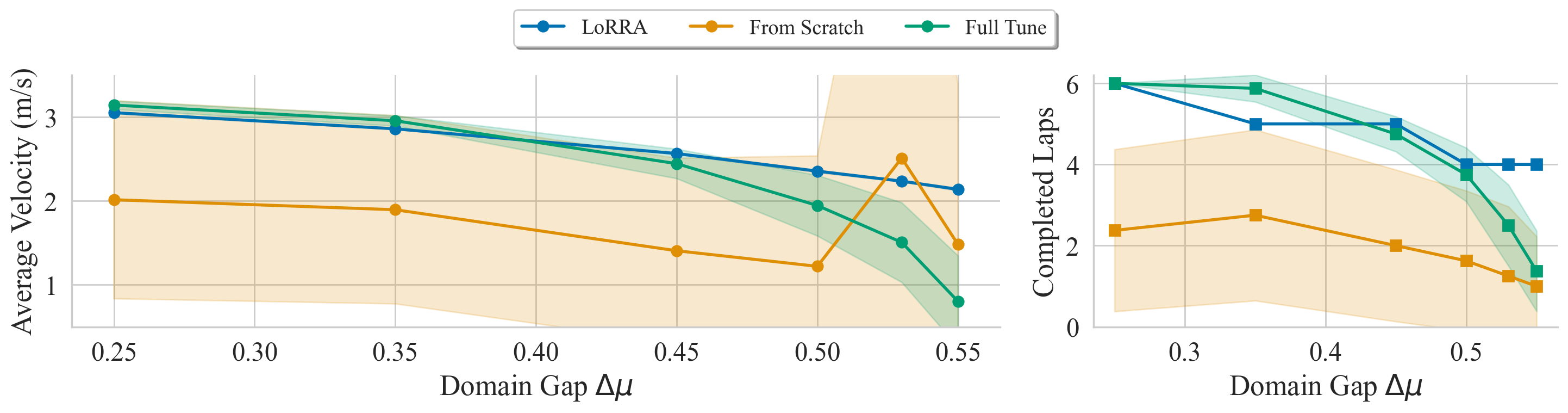}
    \caption{ \footnotesize Performance(Vertical Axis) of RACER using residual dynamics trained with LoRRA(blue), From Scratch(Yellow), Full-Tune(Green) under various domain gaps in terms of friction coefficient(horizontal Axis). \textbf{The larger the domain gaps, the greater the advantages hold by LoRRA.} Shaded Area: variance across seeds. For the definition of each training recipe, see Sec. ~\ref{sec:exp:lowrank}}
    \label{fig:domain_gap_friction}
\end{figure}

\begin{figure}[t]
    \centering
    \includegraphics[width=\linewidth]{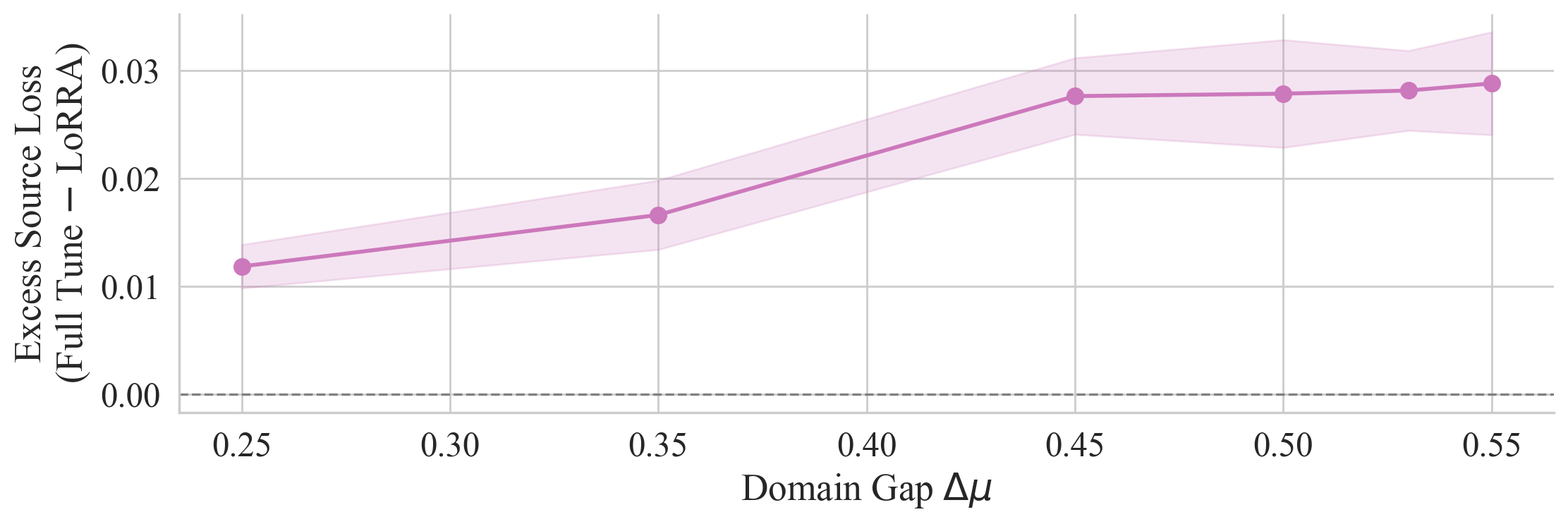}
    \caption{\footnotesize  \textbf{The advantage of LoRRA at preserving source-domain posterior increases with domain gap magnitude.} The horizontal axis shows the domain gap $\Delta\mu$. The vertical axis shows the difference in source-domain loss after adapting residual dynamics to $\mathcal{D}_t$, comparing Full-Tune against LoRRA: $\mathcal{L}(\phi, \mathcal{D}_s)_{\text{Full-Tune}} - \mathcal{L}(\phi, \mathcal{D}_s)_{\text{LoRRA}}$.}
    \label{fig:lorra_train_forgetting_vs_gap}
\end{figure}

\begin{figure*}[t]
    \centering
    \begin{subfigure}[t]{0.45\textwidth}
        \centering
        \includegraphics[width=\linewidth]{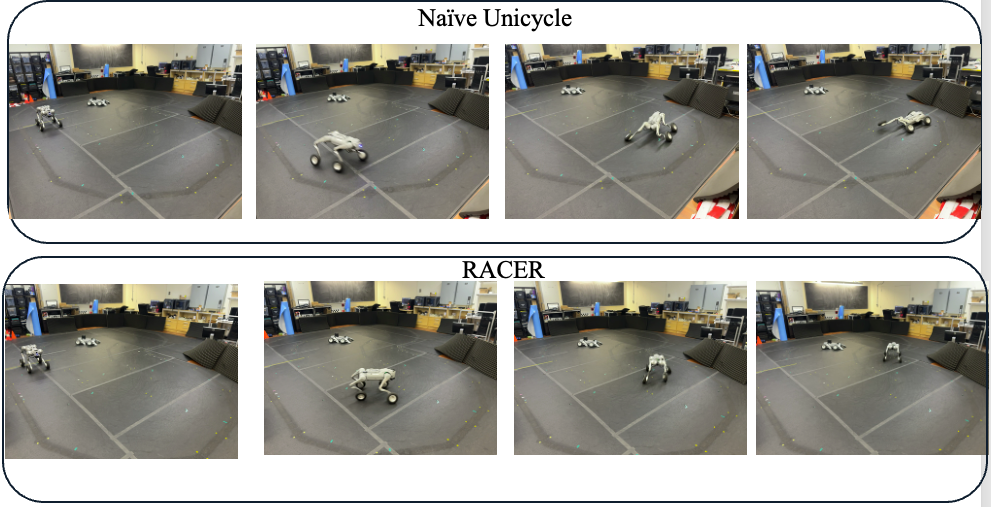}
        \caption{\footnotesize 
        Hardware experiment with velocity cap $3.2$ m/s in
        Sec.~\ref{sec:exp:residual:hardware}.
        \textbf{Naive Unicycle (first row) fails at the first corner,
        while RACER (second row) passes the corner smoothly.}
        }
        \label{fig:vel_ablation_hardware}
    \end{subfigure}
    \hfill
    \begin{subfigure}[t]{0.49\textwidth}
        \centering
        \includegraphics[width=\linewidth]{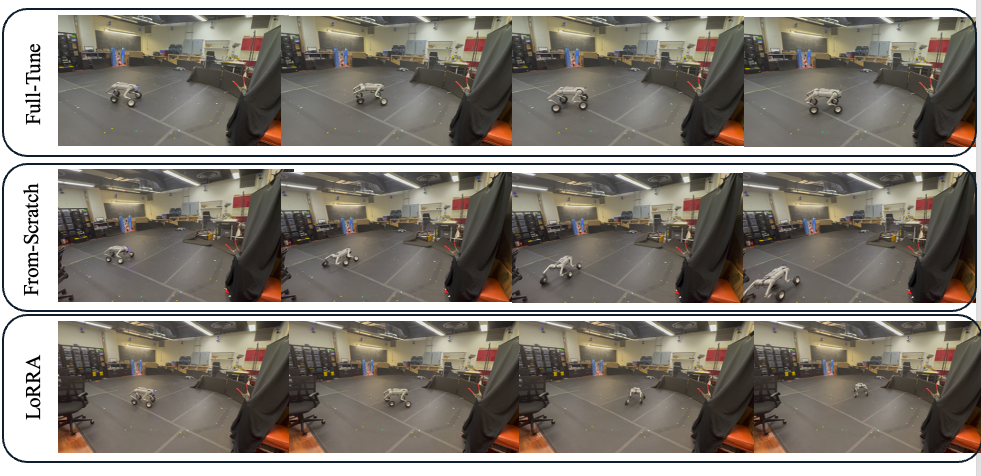}
        \caption{\footnotesize 
        Hardware experiment with joint difference $\Delta K_p=20$ in
        Sec.~\ref{sec:exp:low_rank_hardware}.
        \textbf{Full-Tune (first row) shakes around the first corner due to
        over-fitted residual dynamics; From-Scratch (second row) fails before
        entering the first corner; LoRRA passes the first corner smoothly.}
        }
        \label{fig:LoRRA_ablation_hardware}
    \end{subfigure}

    \caption{\footnotesize Hardware ablation experiments for RACER and LoRRA.}
    \label{fig:hardware_ablation}
\end{figure*}
\subsection{Study of LoRRA}
\label{sec:exp:lowrank}

This section further discusses the training recipe for $g_{\phi}$. We define three strategies for training the residual dynamics:
\begin{itemize}
\item (1) \textbf{LoRRA}, pre-trained on $\mathcal{D}_s$ and fine-tuned under a
low-rank constraint; 

\item (2) \textbf{Full-tune}, pre-trained on $\mathcal{D}_s$ and
fine-tuned without rank constraints; 

\item (3) \textbf{From-scratch}, trained only
on $\mathcal{D}_t$ with no pretraining. This reflects the adaptation methodology of ~\cite{lee2025sym2real} which trains a fresh residual dynamics with target-domain data.
\end{itemize}

For this analysis, the residual dynamics are embedded into the RACER architecture and all previous elements of the control system are the same besides the residual's training method. The comparison between LoRRA and Full-tune validates the role of low-rank constraint; From-scratch validates the significance of dual-phase training. We present two insights.

\begin{insight}[label=ins:dual_phase]
Residual dynamics trained with both source-domain and target-domain data gives racing performance significantly better than the residual dynamics trained with only target-domain data.
\end{insight}

\begin{insight}[label=ins:low_rank]
Low-rank constraint at target-domain adaptation gives higher success rates and higher performance than full-tune and from-scratch when domain gap for ground friction or joint gain is large.
\end{insight}

\subsubsection{Simulation Results}
\label{sec:exp:low_rank_sim}

For the simulation experiment, we define the target domain as a simulation with dynamics that differ from the
source domain. We introduce a controlled gap by changing the tire-ground friction
$\mu_t$ relative to the source $\mu_s = 0.85$, and measure the gap as
$\lvert \mu_s - \mu_t \rvert$. We sweep
$\mu_t \in \{0.30, 0.35, 0.40, 0.45, 0.50, 0.55, 0.60, 0.65\}$, running eight
independent trials per method at each $\mu_t$, each with its own target-domain
data collection and training.

Results are reported in ~\ref{fig:domain_gap_friction}. We present two observations: 
\begin{itemize}

\item Compared to Full-tune and LoRRA, From-Scratch demonstrates significantly worse average performance metrics with much higher cross-seed variance. This is consistent with Insight ~\ref{ins:dual_phase}.

\item Compared to Full-tune, LoRRA's advantage grows with the
domain gap. Under large gaps, LoRRA reaches higher average velocity and presents more stable lap completion than Full-Tune. This is consistent with Insight ~\ref{ins:low_rank}.
\end{itemize}

The key difference between the three training methods is their ability to preserve the posterior of source-domain pre-training. Fig. ~\ref{fig:lorra_train_source_loss_mu0.40} validates this. From-scratch cannot equip the residual dynamics with any knowledge about source-domain by its definition so it shows the highest source-domain loss. Compared to Full-tune, LoRRA has lowest source-domain loss after target-domain adaptation because of low-rank constraint and thus LoRRA is the most capable of preserving source-domain knowledge.

In Fig. ~\ref{fig:lorra_train_forgetting_vs_gap}, we further compare LoRRA and Full-tune's ability of preserving source-domain posterior at different domain gaps. It shows that the advantage of LoRRA over full-tune positively correlates with the magnitude of domain gap, which is consistent with the pattern we observe in Fig. ~\ref{fig:domain_gap_friction} and Insight ~\ref{ins:low_rank}. Similar observations have also been found in the context of LLM . ~\cite{biderman2024loralearnsforgets} finds that fine-tuning LLM with low-rank constraint substantially underperforms Full-Tune on the target in high-rank regimes (code) while nearly matching it in lower-gap regimes (math).

\begin{figure}[t]
    \centering
    \includegraphics[width=\linewidth]{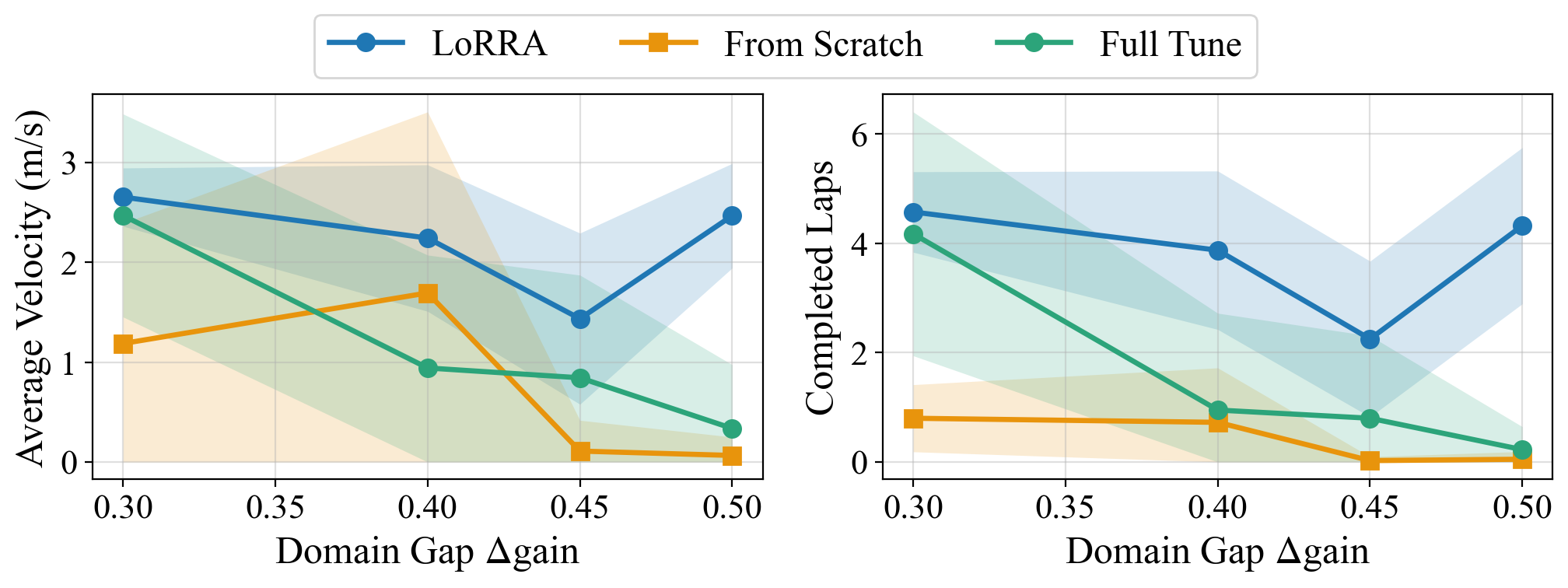}
    \caption{\footnotesize \textbf{LoRRA gives performance consistently greater than or equal to From Scratch and Full Tune under different joint-gain domain gaps}. This figure is the counterpart of Fig. ~\ref{fig:domain_gap_friction} with domain gaps redefined as joint gain difference instead of friction difference.}
    \label{fig:domain_gap_gain}
\end{figure}

To further generalize our results, we repeat the experiment in Fig. ~\ref{fig:domain_gap_friction} by redefining the domain gap as the difference between the joint gains of two domains $K_p$. LoRRA's advantage(and the relationship between such advantage and domain gap) observed earlier holds in Fig. ~\ref{fig:domain_gap_gain}, but in a much noisier way. The main reason is that domain gap in terms of joint gain is more likely to trigger immediate failure of the policy (falling down to the ground) while domain gap in terms of ground-friction results in smoother dynamics difference.

In conclusion, when the size of target-domain data is essentially small, leveraging source-domain posterior gives us stronger racing performance compared with exclusively using target-domain data(Insight ~\ref{ins:dual_phase}). When the domain-gap is large, low-rank constraint can help better preserve source-domain posterior and gives better racing performance compared with Full-tune(Insight ~\ref{ins:low_rank}).

\subsubsection{Hardware Results}
\label{sec:exp:low_rank_hardware}

\begin{table}[t]
\centering
\footnotesize
\setlength{\tabcolsep}{6pt}
\renewcommand{\arraystretch}{1.15}
\caption{\footnotesize \textbf{Hardware LoRRA ablation (counterpart of Fig.~\ref{fig:domain_gap_gain}).} Racing metrics (mean~$\pm$~std over three independently trained seeds) for the three residual-training recipes under an increasing joint-gain domain gap $\Delta K_p = |K_{p,s} - K_{p,t}|$ (larger $\Delta K_p$ $=$ larger gap; the residual is pre-trained at the nominal gain $K_{p,s}=70$). \textbf{LoRRA completes laps robustly across all gaps, whereas Full-Tune degrades as the gap grows and fails entirely at $\Delta K_p=50$, and From-Scratch fails at nearly every gap.} Velocity is averaged over all seeds (a failed run contributes $0$ velocity); \textbf{bold} marks the best value per metric.}
\label{tab:hardware_lorra}
\begin{tabular}{@{}c l c c@{}}
\toprule
$\Delta K_p$ & \multirow{2}{*}{Method} & Clean laps & Avg. vel. \\
(gap) & & (max is 3) & (m/s) \\
\midrule
\multirow{3}{*}{0}
 & LoRRA        & $\mathbf{3.00 \pm 0.00}$ & $1.94 \pm 0.07$ \\
 & Full-Tune    & $2.67 \pm 0.47$          & $\mathbf{2.06 \pm 0.05}$ \\
 & From-Scratch & $0.33 \pm 0.47$          & $0.59 \pm 0.83$ \\
\midrule
\multirow{3}{*}{30}
 & LoRRA        & $\mathbf{2.33 \pm 0.94}$ & $\mathbf{1.61 \pm 0.08}$ \\
 & Full-Tune    & $1.00 \pm 0.82$          & $0.60 \pm 0.44$ \\
 & From-Scratch & $0.00 \pm 0.00$          & $0.00 \pm 0.00$ \\
\midrule
\multirow{3}{*}{50}
 & LoRRA        & $\mathbf{2.33 \pm 0.94}$ & $\mathbf{1.29 \pm 0.10}$ \\
 & Full-Tune    & $0.00 \pm 0.00$          & $0.00 \pm 0.00$ \\
 & From-Scratch & $1.00 \pm 1.41$          & $0.80 \pm 1.14$ \\
\bottomrule
\end{tabular}
\end{table}

On hardware, we induce a controlled domain gap by varying the low-level joint tracking gain on our hardware, using the same joint-gain gap definition as the simulation study in Fig.~\ref{fig:domain_gap_gain}. The residual is pre-trained with simulation data collected at the nominal gain $K_{p,s}=70$, and we quantify the sim-to-real gap as $\Delta K_p = |K_{p,s} - K_{p,t}|$ between the simulation joint gain $K_{p,s}$ and hardware joint gain $K_{p, t}$. We test each recipe with $\Delta K_p \in \{0, 30, 50\}$. A larger $\Delta K_p$ means larger sim-to-real gap. For each $\Delta K_p$, we collect a hardware trajectory and run three independently-seeded train per recipe using the same hardware dataset. We report the across-seed mean and variance per recipe and per $\Delta K_p$ in Table~\ref{tab:hardware_lorra}.

The results reproduce both insights on hardware. From-Scratch, which uses only target-domain data, fails to complete laps at almost every gap and shows the highest across-seed variance, confirming the value of dual-phase training (Insight~\ref{ins:dual_phase}). Among the two dual-phase recipes, LoRRA and Full-Tune are comparable at zero gap ($\Delta K_p=0$), with Full-Tune even slightly faster. As the gap grows, however, Full-Tune degrades sharply while LoRRA keeps completing laps at every gap. This is consistent with Insight~\ref{ins:low_rank}.

\section{Conclusion}
\label{sec:conc}
We presented RACER, a hierarchical control framework for autonomous quadruped racing. We exhaustively validated two main ingredients of our pipeline: residual dynamics and LoRRA. We identify two potential questions for future research. (A) Can hardware adaptation of the residual dynamics happen on-fly safely? (B) How to adapt residual dynamics in wild environments? 

\section*{Acknowledgment}
We appreciate the intellectual support and technical advice from Shengfan Cao. We appreciate the photo-filming support from Shengfan Cao and Simon He.
\balance
\bibliographystyle{IEEEtranN}
\bibliography{references}
\appendix


\subsection{DIAL-MPC Cost Function}
\label{app:mppi:global}

The DIAL-MPC planner optimizes trajectories to stay closely to a globally optimum reference trajectory computed offline. When a reference trajectory
$\bigl(x^{\star}(s),\, y^{\star}(s),\, \psi^{\star}(s),\, v^{\star}(s)\bigr)$
parameterised by arclength $s$ is available, the contouring term is
replaced by an explicit trajectory-tracking error. For each rollout
state $s^{(k)}_h = (x_h^{(k)}, y_h^{(k)}, \psi_h^{(k)}, v_{x,h}^{(k)}, \omega_{z,h}^{(k)})$,
we project $(x_h^{(k)}, y_h^{(k)})$ to its closest arclength using the
same grid projector, query the reference at that arclength, and sum
\begin{equation}
  J_{\mathrm{track}}^{(k)}
  \;=\;
  \lambda_c \sum_{h=0}^{H} \bigl(\Delta x_h^2 + \Delta y_h^2 + \Delta \psi_h^2\bigr)
  \;+\;
  \lambda_v \sum_{h=0}^{H} \Delta v_{x,h}^2,
  \label{eq:global_track_cost}
\end{equation}
with $\Delta x_h = x^{\star}(s_h^{(k)}) - x_h^{(k)}$ and analogous
definitions for $\Delta y_h, \Delta \psi_h, \Delta v_{x,h}$. The
reference trajectory itself is precomputed offline by solving a
nonlinear program on a dynamic-bicycle model of the Go2-W in CasADi with  velocity cap $v_x \leq c$ (Sec. ~\ref{sec:exp:residual:sim}). The maximum deviation metrics in Sec. ~\ref{sec:exp} is computed relative to this reference trajectory.

\subsection{Details on LoRRA}
\label{App:LoRRA_details}
Denote the parameter matrix of $g_{\phi}$'s $l$th layer as $W_l \in \mathbb{R}^{d \times k}$. The update rule is parameterized as:

\begin{equation}
\label{eq:LoRRA_update}
W_l := W_l^0 \; + \; \frac{\alpha}{r}BA 
\end{equation}
where $W_l^0$ is the pretrained weights from source-domain which is frozen throughout target-domain adaptation. $B \in \mathbb{R}^{d \times r}$ and $A \in \mathbb{R}^{r \times k}$ can be updated throughout the back-propagation passes of target-domain adaptation. $r \le \min(d,k)$ is the adaptable rank.

Two hyperparameters are of our interest: $r$ and $\alpha$. $r$ decides the degree of freedom allowed for adaptation, and $\alpha$ decides the magnitudes of the update.

To simplify our hyperparameter tuning, we assume each layer has the same $r$ and $\alpha$ value. Following LLM community's common practice of hyper-parameter tuning ~\cite{hu2022lora, biderman2024loralearnsforgets}, $\alpha$ is fixed to be $0.35$ and we sweep through $r$ in the candidates $r \in \{2, 4, 6, 8, 10, 12\}$ and pick the one that can give the best performance for sim-to-sim domain adaptation with ground-friction domain gap $\Delta \mu = 0.55$. We found $r=6$ to give the best performance. The hyper-parameter pair $r =6$ and $\alpha = 0.35$ is used throughout for all the sim-to-sim and sim-to-real adaptations in ~\ref{sec:exp}. Under this setting, the number parameters being updated are $17.7$ percent of the total parameters of $g_{\phi}$.


\end{document}